 \documentclass[final,1p,times]{elsarticle}

\usepackage{booktabs}
\usepackage{natbib}
\usepackage{multirow}
\usepackage{amsfonts}
\usepackage{amssymb}
\usepackage{amsmath}
\usepackage[usenames,dvipsnames]{pstricks}
\usepackage{lscape}
\usepackage{rotating}
\usepackage{hyperref}
\usepackage{enumitem}
\usepackage{multirow}
\usepackage{color}
\usepackage{lscape}
\usepackage{colortbl}
\usepackage{url}
\usepackage{algorithm}
\usepackage{algorithmic}
\usepackage{lineno}

\usepackage{setspace} 
\newcommand{\mytitle}{Detecting animals in African Savanna with UAVs and the crowds}

\journal{Remote Sensing of Environment}

\begin{document}
\begin{frontmatter}

\title{\mytitle}
\author[1]{Nicolas Rey}
\author[2]{Michele Volpi}
\author[1,3]{St{\'ephane Joost}}
\author[2,4]{Devis Tuia}
\cortext[MPgrant]{Corresponding Author: Devis Tuia, devis.tuia@wur.nl.}

\address[1]{Laboratory of Geographical Information Systems (LASIG), School of Architecture, Civil and Environmental Engineering (ENAC), Ecole Polytechnique F{\'e}d{\'e}rale de Lausanne (EPFL), Switzerland}
\address[2]{MultiModal Remote Sensing, Department of Geography, University of Zurich, Switzerland}
\address[3]{The Savmap Consortium}
\address[4]{Laboratory of Geo-Information Science and Remote Sensing, Wageningen University \& Research, the Netherlands}

\begin{abstract}
\noindent \textbf{This is the pre-acceptance version, to read the final version published in Remote Sensing of Environment, please go to: \href{https://doi.org/10.1016/j.rse.2017.08.026}{10.1016/j.rse.2017.08.026}}.

Unmanned aerial vehicles (UAVs) offer new opportunities for wildlife monitoring, with several advantages over traditional field-based methods. They have readily been used to count birds, marine mammals and large herbivores in different environments, tasks which are routinely performed through manual counting in large collections of images.
In this paper, we propose a semi-automatic system able to detect large mammals in semi-arid Savanna. It relies on an animal-detection system based on machine learning, trained with crowd-sourced annotations provided by volunteers who manually interpreted sub-decimeter resolution color images. The system achieves a high recall rate and a human operator can then eliminate false detections with limited effort. Our system provides good perspectives for the development of data-driven management practices in wildlife conservation. It shows that the detection of large mammals in semi-arid Savanna can be approached by processing data provided by standard RGB cameras mounted on affordable fixed wings UAVs. \newline

\end{abstract}


\end{frontmatter}

%\linenumbers

\section{Introduction}

In the fragile ecosystems of semi-arid Savanna, any change in the equilibrium between precipitation, grazing pressure and bush fires can lead to long-term land degradation, such as the reduction in grass cover and bush encroachment \citep{trodd_monitoring_1998}. To avoid overgrazing, the populations of grazers must be kept in adequacy with the grass availability, which is subject to meteorological conditions. For this purpose, land managers need to regularly estimate the amount of wildlife present on their territory. Thus, monitoring wildlife populations is crucial towards conservation in wildlife farms and parks.

To carry out wildlife censuses, traditional methods include transect counts on land or from a helicopter, and camera traps. While a total count is usually not possible over large areas, these methods estimate the population density based on observations localized along a predefined path (see \citep{alienor2017,aebischer2017} and references therein). These methods are expensive {(e.g. in the case of the Kuzikus reserve considered in this paper, helicopter costs for a single survey are between 1000\$ and 2500\$)}, require trained human experts to screen large amounts of data and are consequently not suitable for regular censuses over large areas.

In recent years, unmanned aerial vehicles (UAVs) have been used to detect and count wildlife such as birds, marine mammals, and large herbivores \citep{linchant_are_2015}. Compared to traditional methods, UAVs offer several advantages: they cover large areas in a short amount of time and can be used in inaccessible and remote areas, yet they are cheaper and easier to deploy than helicopters. Moreover, they are safer for the pilot, who can stay on the ground and avoiding retaliations from poachers.

However, UAVs collect large amounts of color images with sub-meter to sub-decimeter spatial resolution, of which only few contain animals. Furthermore, the animals cover only an infinitesimal area of the images and their color might blend in smoothly with background vegetation and soil. Therefore, identifying and counting single animals across large collections of images is extremely complex and time-consuming, preventing land managers from using UAVs on a regular basis. 

Despite these challenges, recent developments in object detection pipelines in both computer vision~\citep{malisiewicz2011iccv,girshick2014cvpr} and remote sensing~\citep{Tue13,Mor14,Akc16}, provide promising techniques to {semi-}automatically localize and count animals. {We refer to these methods as semi-automated and not as fully automated since they rely on supervised learning paradigms, thus requiring annotated ground truth to be trained.} Still, as the human effort required to make sense of the aerial images is reduced, the overall benefits of using UAVs are significantly increased.

The use of UAVs in wildlife monitoring and conservation is well documented (e.g. \citet{linchant_are_2015}), but only few studies have implemented {semi-}automatic detection pipelines. 
\citet{dorffer_uas-based_2013} proposes to detect seagulls by combining supervised classification of RGB images with geometric rules. \citet{kudo2012cost} present a pipeline to count salmons in aerial images using simple color thresholding after contrast adjustment. Such approaches are only possible if the animals are visually very similar and exhibit distinctive colors that contrast with the background.
\citet{chabot_evaluation_2012} detect geese by manual counting single animals in UAV images. \citet{maire_convolutional_2014} adopt more advanced machine learning tools for the detection of dugongs. They obtain promising results by training a deep neural network and address the problem of scarcity of training samples by replicating them through random rotations and scalings applied to confident missclassifications (a technique related to hard negative mining~\citep{malisiewicz2011iccv}).

In this paper, we present a data-driven machine learning system for the {semi-}automatic detection of large mammals in the Savanna ecosystem characterized by complex land-cover. We perform animal detection on a set of sub-decimeter resolution images acquired in the Namibian Kalahari desert and train our system using animals annotated by digital volunteers using the Micromappers crowdsourcing platform \citep{ofli_combining_2016}. {We focused on large mammals for two main reasons: first, they stood out compared to the background, while smaller animals such as meerkats are not clearly visible and could be too easily confused with rocks or bushes by the volunteers. Secondly, larger animals also mean more pixels available to learn the appearance of the animals' furs, which leads to less signal mixing, to more discriminative features and to a more accurate system overall.} We show that the system achieves high recall rate, and high overall accuracy can be obtained if a human operator can verify the detections, reduce the false positives and verify true negatives, and retrain the detector. {This last technique, known as active learning~\citep{tuia2011jstsp}, aims at focusing the operator's effort on instances with low detection confidence and its benefits are shown by our experimental results, where only 1h was required to correct the crowd-sourced dataset of several errors (mainly animals missed by the volunteers).}
{The main contributions of the paper are:}

\begin{itemize}
\item[-] {A pipeline for semi-automatic animal detection in semi-arid Savanna that uses affordable UAV platforms with off-the-shelf RGB cameras;}
\item[-] {A complete study of the model's parameters to provide intuitions about the trade-offs between acquisition settings, image resolution and the complexity of the appearance descriptors involved;}
\item[-] {A discussion of the promising performances of the system in a real deployment scenario in the Kuzikus reserve in Namibia, including the quasi real time improvement of the model.}
\end{itemize}

\section{Study area and data } \label{sec:data sets}

\subsection{The Kuzikus wildlife reserve}

Kuzikus is a private wildlife reserve that covers 103 km$^2$ (10'300 hectares), located on the edge of the Kalahari in Namibia. The Kalahari is a semi-arid sandy Savanna that extends across Botswana, South Africa and Namibia. It is home of an important variety of animals, including many large mammal species \citep{kuzikus_species}. About 3'000 individuals from more than 20 species populate the reserve, including Common Elands (\emph{Taurotragus oryx}), Greater Kudus (\emph{Tragelaphus strepsiceros}), Gemsboks (\emph{Oryx gazella}), Hartebeests (\emph{Alcelaphus buselaphus}), Gnus (\emph{Connochaetes gnou} and \emph{Connochaetes taurinus}), Blesboks (\emph{Damaliscus albifrons}), Springboks (\emph{Antidorcas marsupialis}), Steenboks (\emph{Raphicerus campestris}), Common Duickers (\emph{Sylvicapra grimmia}), Impalas (\emph{Aepyceros melampus}), Burchell's Zebras (\emph{Equus quagga burchellii}), Ostriches (\emph{Struthio camelus australis}) and Giraffes (\emph{Giraffa camelopardalis giraffa}).

\subsection{The SAVMAP 2014 UAV campaign} \label{ssec:SAVMAP_campaign}

\begin{figure}[!t]
\centering
\includegraphics[width=0.8\textwidth]{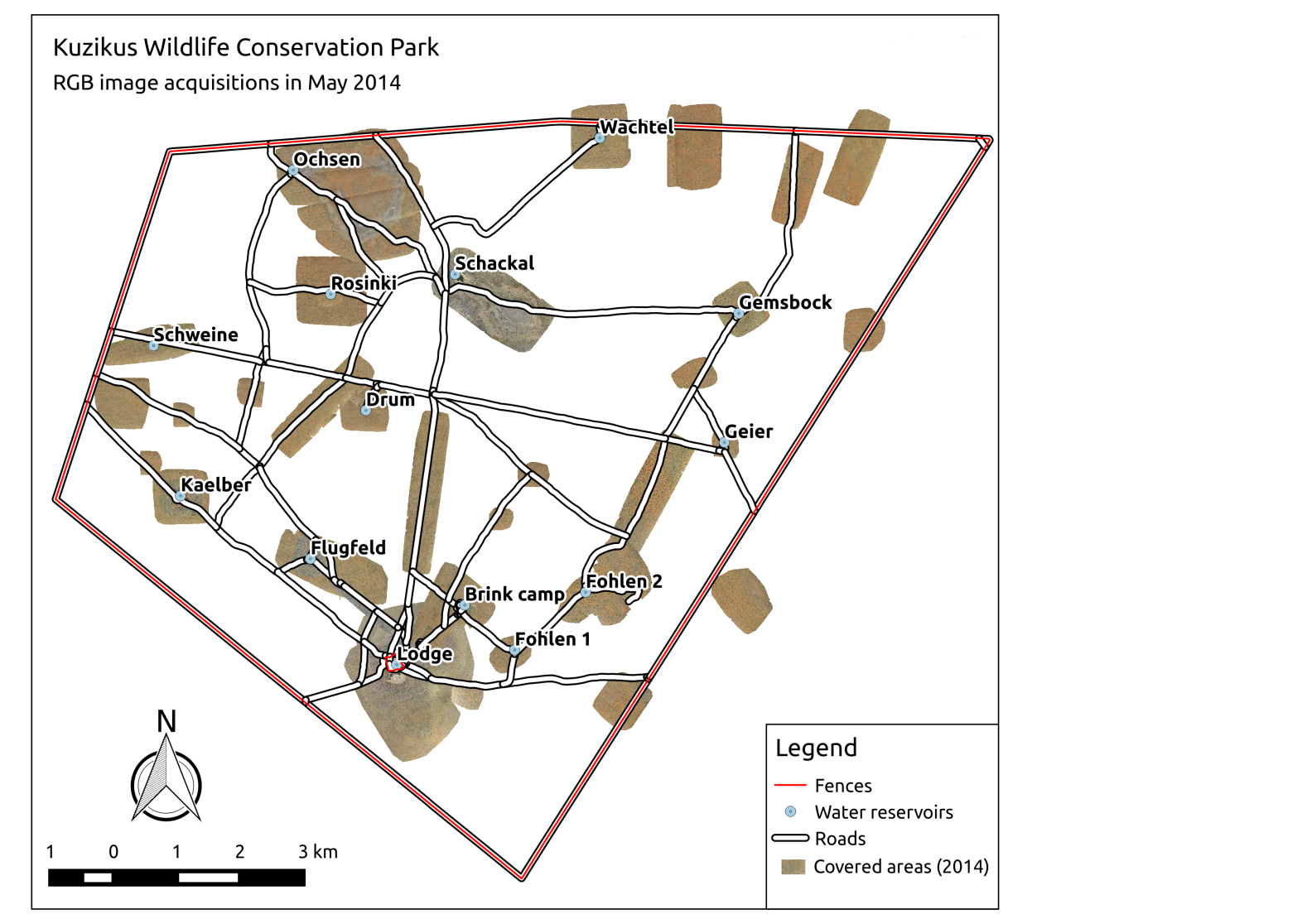}
\caption{Map of the Kuzikus Wildlife Conservation Park and areas covered by the 2014 RGB dataset.\label{fig:Kuz}}
\end{figure}

The SAVMAP Consortium (\url{http://lasig.epfl.ch/savmap}) acquired a large aerial image dataset during a two-week campaign in May 2014. {It is composed of five flights, between May 12 and May 15, 2014.} The images were acquired with a Canon PowerShot S110 compact camera mounted on an eBee, a light UAV commercialized by SenseFly (\url{https://www.sensefly.com}). Each image is $3000 \times 4000$ pixels in size and comprises three bands in the Red Green and Blue (RGB) domains, with a radiometric resolution of 24 bits. The ground sampling distance is approximately 4~cm per pixel. The extent of the reserve mapped by the 2014 SAVMAP campaign is illustrated in Fig.~\ref{fig:Kuz}.

\subsection{Animals annotation via crowd sourcing}

In order to obtain a ground truth of the position of all large animals, a crowd-sourcing campaign was set by MicroMappers (\url{https://micromappers.wordpress.com/}). {A total of 232 digital volunteers participated in the operation}. The volunteers were asked to draw a polygon around each animal they detected in the images, without distinction between species. They did not have to report signs of animal presence, such as Aardwolves' holes or termite mounds. {Each image is inspected by at least three volunteers, with a maximum of ten. On average, the images were seen by five volunteers~\citep{ofli_combining_2016}}.

\begin{figure}[!t]
\centering
\includegraphics[width=.99\linewidth]{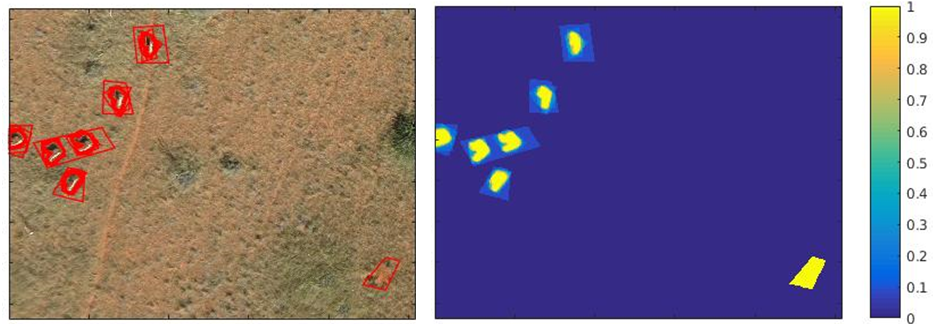}
\caption{Example of the crowdsourced image annotations. Left: annotations of the volunteers, represented as red polygons (one polygon per annotation and user). Right: Annotation confidence map. All the areas in green-to-yellow are retained as the ground truth annotations, with the exception of the one in the bottom right, since it was annotated only by one volunteer.\label{fig:conf}}
\end{figure}

The volunteers visually analyzed 6'500 images and drew 7'474 polygons in 654 images containing animals. After merging the overlapping polygons and removing objects tagged only by a single volunteer (as the bottom right annotation in Fig.~\ref{fig:conf}), 976 annotations were kept. {Since the number of volunteers per image varied between three and ten, we used as ground truth the surface that was tagged as ``animal'' by at least half of the annotators who considered it (areas in green-to-yellow colors in the right panel Fig.~\ref{fig:conf}).} To avoid false annotations, we visually inspected them to confirm or infirm animals presence. It took 30 minutes to verify the 976 annotations, leading to the removal of 21 spurious ones. More details on the acquisition of annotations can be found in~\citet{ofli_combining_2016}. Note that the same animals could be observed in several consecutive, overlapping images. This effect is beneficial when training appearance models, since the different angles and poses characterizing animals better cover the appearance variability of the class of interest.
However, note that the current system has no tracking component (nor ambition to track animals), so it cannot detect if a same animal has been detected multiple times during the same flight.
{This means that when detecting animals in new images, there is a potential risk of counting a same animal multiple times. However, due to the scarcity of animals in images overall and since the task can only by definition lead to an approximation of the real animal number, there is still a clear advantage for using the proposed system as compared with traditional techniques.}

\section{Methods} \label{sec:methods}

In the following, we present the main components of our machine learning pipeline, as well as the iterative refinement with active learning.

\subsection{Animals detection system}\label{sec:objd}

\begin{figure}[!t]
\centering
\includegraphics[width=10cm]{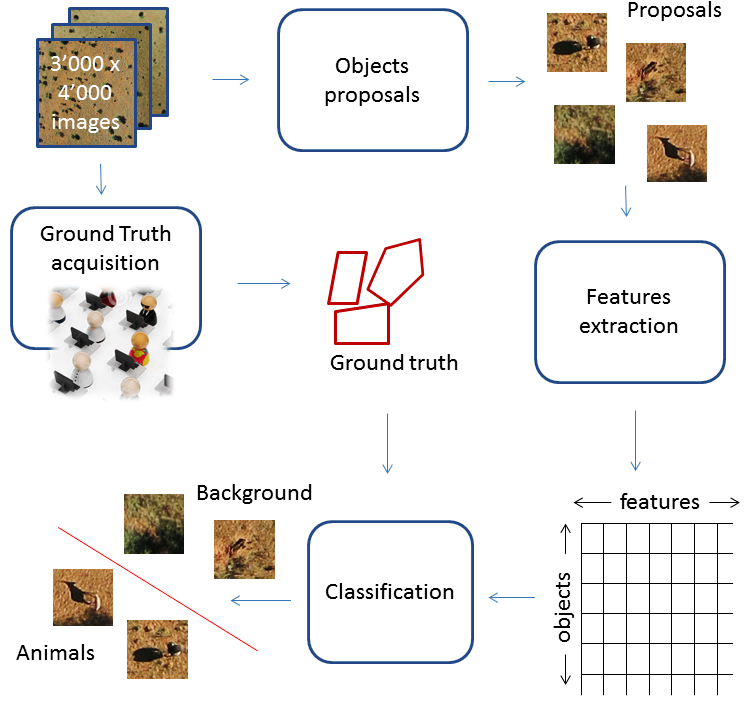}
\caption{Pipeline for the  detection of animals by object proposals}\label{fig:detection_pipeline}
\end{figure}

The proposed system is composed of three steps, as illustrated in Fig.~\ref{fig:detection_pipeline}: 

\begin{enumerate}
\item[1)] Definition of \emph{object proposals}~\citep{alexe2012tpami,Vol16}, i.e.  regions of interest, which are likely to contain an animal {(Section~\ref{ssec:op})}; 
\item[2)] Extraction of a set of mid-level appearance descriptors, or \emph{features}, defining animals meaningful visual characteristics {(Section~\ref{ssec:features})}; 
\item[3)] A classification model, or \emph{detector}, learning from the training set proposals and their features to detect animals in new regions {(Section~\ref{ssec:classifier})}.
\end{enumerate}

\subsubsection{Object proposals} \label{ssec:op}

Na\"ive approaches to object detection require a classifier to scan all possible windows centered on each pixel, at every possible scale. {Although exhaustive, this strategy is computationally heavy}, since windows containing animals correspond to a very small fraction of the image data and most computations are wasted. {One could} discard regions of the image where the object is unlikely to occur, {for instance by modeling class co-occurrences and discarding unlikely backgrounds, as~\citet{Mor15} did for the detection of cars. However, this strategy is hardly applicable to a semi-arid Savanna where animals can stand everywhere and co-occurrence of animals to background is uniform.} 

The alternative solution {proposed} is to quickly find {an overcomplete set of} regions likely to contain objects of interest and to consider them as candidates to be screened. In computer vision this concepts is known as \emph{object proposals} generation. Object proposals have long been used in object detection pipelines in natural images \citep{zitnick2014eccv,alexe2012tpami,uijlings2013ijcv}. The aim of object proposals {is to provide meaningful, context-dependent and adaptive spatial support from which it is possible to extract meaningful and informative appearance descriptors needed to train an accurate object detector, while reducing the prediction space of the latter (see Section~\ref{ssec:classifier})}. This is because we process only a much smaller candidate subset of all the possible image windows, which are likely to contain an animal. This step mainly discards regions that are very likely to \emph{not} contain any positive instance. Such a subset can be defined, for example, as the ensemble of windows containing high density of closed contours \citep{zitnick2014eccv}, as a set of windows containing object-like color and edge distribution \citep{alexe2012tpami} or as windows containing sets of similar regions dissimilar from those not contained in the bounding box \citep{uijlings2013ijcv}. 

A good generator of object proposals must lead to high recall rates, i.e. it must {cover with proposals} all the positive objects of interest{. It has usually low precision, because many ambiguous proposal windows that do not contain any object of interest are also included (overcomplete set). Such a trade-off is acknowledged, since the detection of the animals is left to the detector.} %The proposal generation step is mainly aimed at reducing the search space by discarding regions \emph{not} containing positive instances with high-probability.} 

Our object proposal system relies on two observations: 
\begin{itemize}
\item[-] Standing animals cast a shadow. {We define proposal based on a thresholding of the value channel issued by the HSV transform.} Connected regions with an area smaller than three pixels are discarded and the centroids of the remaining regions define the proposals. 
\item[-] Laying animals, as well as animals located in the shade of a tree do not cast a distinctive shadow. To cope with this, {we group responses of a Sobel edge detector applied on the blue channel. The filter produces a map of edge scores which is binarized by a threshold. The centroids of the connected regions larger than three pixels define the edge-based proposals. This second approach proved to be very informative, because of the high contrast and sharp edges of animal furs.}
\end{itemize}

Either method defines a set of proposals, but the highest recall is obtained by combining the two. For most animals the two methods produce proposals in agreement, i.e. very similar to each other, both in location and size{, leading to duplicate proposals.} To avoid this problem, we merged proposals closer than 15 pixels (i.e. closer than 60 cm). {This threshold is in principle smaller than the average distance between close-by animals, and also corresponds to the average width of an animal in our dataset.} {Finally,} all images are downgraded to 8~cm resolution (i.e. by a factor 2), since we observed that the results did not change significantly, but computational effort was greatly reduced: {after such downgrading, the number of pixels is reduced by 4, and consequently the computation of all the features is reduced by a proportional amount. The efficiency of the rest of the proposed system is not affected by the change in resolution, as it scales quadratically to the number of training examples and only linearly with dimensionality, which depends on the type of appearance descriptor. Note that the dimensionality of the appearance features does not depends on the spatial resolution, but only their computation is affected. We will detail this observation on the resolution in the experimental section.}

\subsubsection{Features} \label{ssec:features}

To train the detector, each proposal must be represented by features describing its appearance (e.g. colors and textures). Since our detector is based on a linear classifier (as detailed in Section~\ref{ssec:classifier}), we want these features to translate complex and nonlinear appearance variations into linearly separable characteristics. Furthermore, we want to employ features that are invariant to the rotation of the window containing the animal, since absolute orientation must not affect the detection scores. In this work, we considered two types of features:
\begin{itemize}
\item[-] \emph{Histogram of colors} (HOC): This descriptor summarizes the probability distribution of colors in a given image patch, by computing their histogram. It is computed over a square region of $25 \times 25$ pixels centered on the proposal. The values are summarized in 10 bins and, for each proposal, the histograms of the red, blue and green bands are concatenated, yielding a 30-dimensional feature vector.
\item[-] \emph{Bag-of-visual-words} (BoVW,~\citet{Siv03}): like color histograms, BoVW relies on a quantization of the image data. However, rather than binning color channels independently, BoVW accounts for dependencies across the whole RGB space. Our BoVW extraction pipeline is as follows {(Fig.~\ref{fig:BOVW})}:

\begin{figure}[!t]
\includegraphics[width = \linewidth]{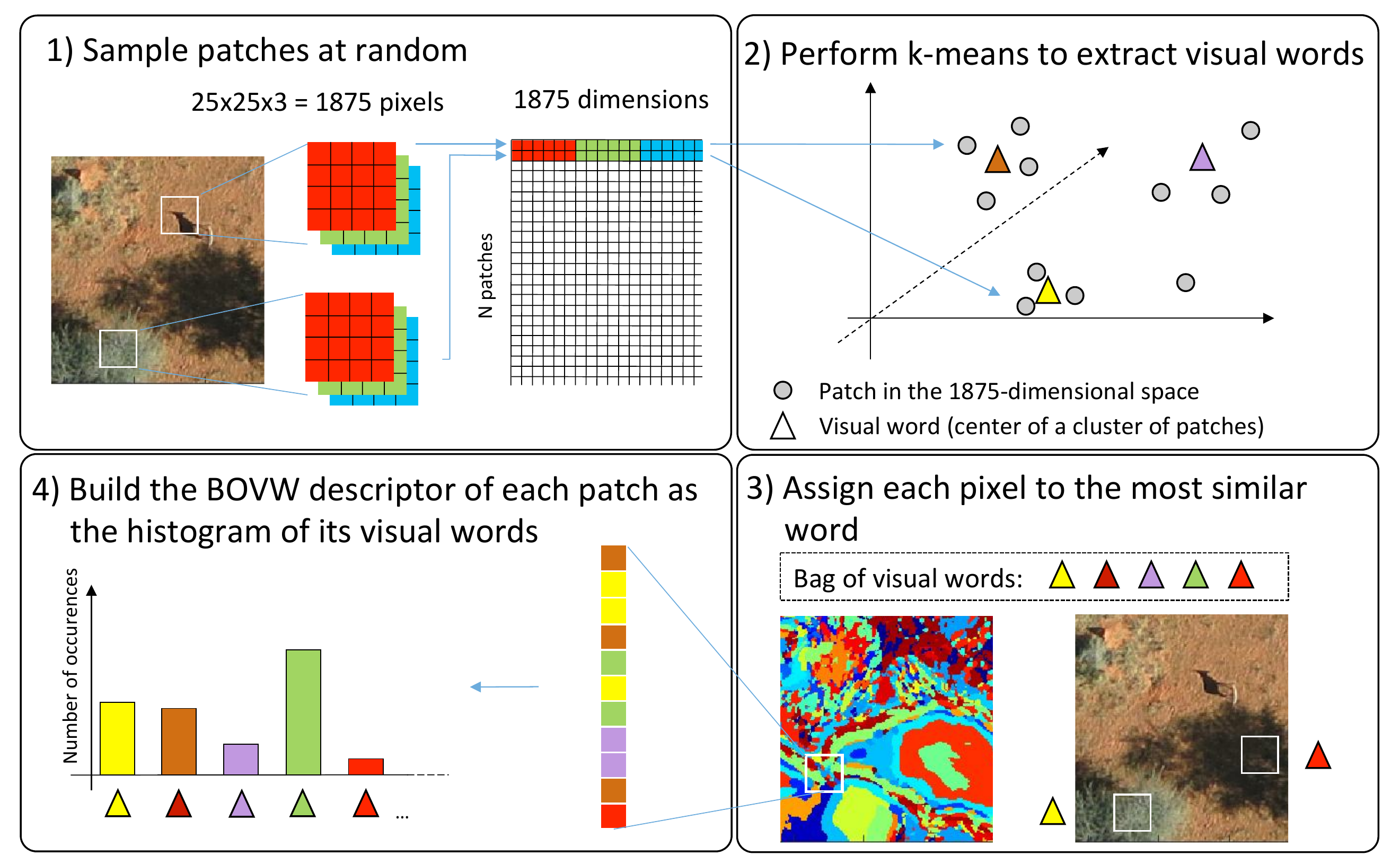}
\caption{Extraction of the BoVW descriptor in four steps. \label{fig:BOVW}}
\end{figure}

\begin{enumerate}
\item We extract 20'000 $25 \times 25$ pixels patches from 100 different images. To ensure that animals are well represented, we enforce that 5'000 samples are located on animals annotations, while the rest is sampled randomly across background regions. {For each window, we then concatenate the RGB pixel values in a single 1'875-dimensional vector (i.e. $25 \times 25 \times 3$)}.
\item We apply the $k$-means clustering algorithm to group the feature vectors into $k$ clusters. {The centers of the clusters are used} as the representative patterns in the images, or \emph{visual words}. 
\item All the possible $25 \times 25 \times3$ patches in all the images are then assigned to the closest among the $k$ visual words, to generate a dense map of visual words. 
\item {Like} the HOC features above, a BoVW feature vector {is a $k$-dimensional} histogram of visual words occurring in the $25 \times 25$ window surrounding the candidate pixel {considered}.
\end{enumerate}
\end{itemize}

The BoVW procedure offers a series of beneficial aspects over pure color-based descriptors. First, the binning of the image is more semantic, as the presence of a given visual word corresponds to the occurrence of a specific pattern in the window. Secondly, mapping the images into a space extrapolated from overlapping windows ensures that descriptive features are spatially smooth, which is a prior belief on image data, while locally keeping signals variance. 

\subsubsection{Detector: ensemble of Exemplar SVM} \label{ssec:classifier}

\begin{figure}[!t]
\centering
\includegraphics[width=12cm]{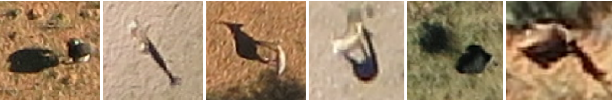}
\caption{Visual heterogeneity within the positive class (\textit{animals}).}\label{fig:visual heterogeneity}
\end{figure}

Once the proposals have been defined and features extracted, we train the animal detector on these inputs. The detection task is formulated as a binary classification problem, involving a positive (``animal'') and negative (``background'') class. {The problem is challenging for two main reasons:}
\begin{itemize}
\item[-] Both classes are very heterogeneous, as shown in Fig.~\ref{fig:visual heterogeneity}. On the one hand, most animals have a light fur, but one can also find darker, brown and gray/black (Ostriches) individuals. {Shapes vary strongly and a projected shadow is a frequent, but not persistent characteristic.} On the other hand, the background class contains diverse land-cover types such as bare soil, sand, roads, annual and perennial grasses, with sparse shrubs and trees and the corresponding shadows. Aardwolves holes are frequent and appear as black spots that are visually very similar to animals' shadows. {Beside the complexity of the classification task, many background objects can be confused with animals.}
\item[-] Animals are very rare in terms of total number of instances and occupy only a tiny fraction of the images in terms of spatial coverage. Depending on the local animal density and on the land cover type, which influences the amount of object candidates, the ratio of positive to negative proposals varies between 1:200 and 1:500, while the total area occupied by an animal and its shadow is around 5'000 pixels (roughly 8~m$^2$), which is only $0.04$\% of an image.
\end{itemize}

\begin{figure}[!t]
\begin{tabular}{ccc}
\includegraphics[width= .45\linewidth]{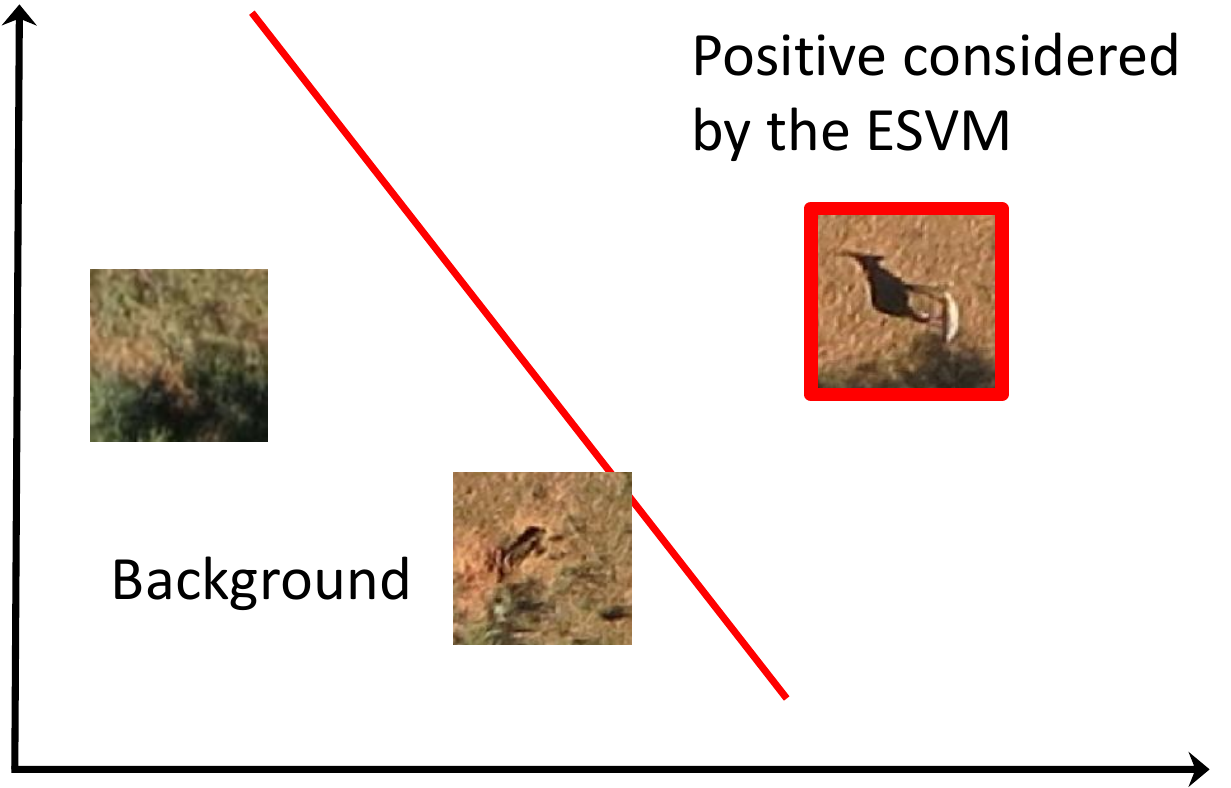}& &
\includegraphics[width= .45\linewidth]{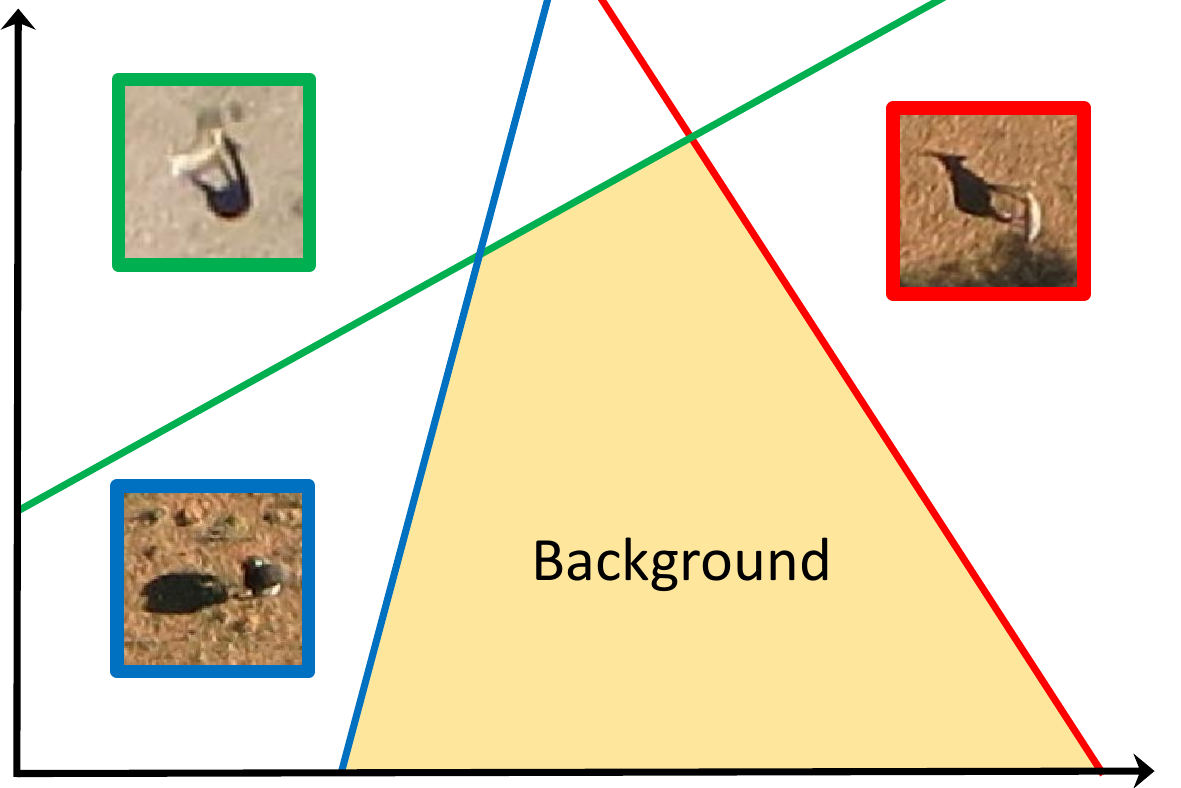}\\
(a) && (b) \\
\end{tabular}
\caption{a) An exemplar SVM separating a single positive example (framed in red) against a negatives set made of background and other positive examples. b) The final decision function is an ensemble of linear models, one per positive example. Each color corresponds to a single ESVM; the orange area indicates the background class prediction.}\label{fig:ESVM}
\end{figure}

To tackle these issues we adopt the Ensemble of Exemplar Support Vector Machines (EESVM) detector~\citet{malisiewicz2011iccv}. The EESVM is composed by an ensemble, where each member learns a separate model for each positive instance in the training set, rather than learning a global model at once. Each model is a binary SVM trained to discriminate between a single positive and many negative instances{, and is known as an} \emph{Exemplar SVM}, see Fig.~\ref{fig:ESVM}a. {This strategy offers flexibility to encode very diverse positive examples, while keeping the overall robustness of a single detector given by the ensemble learning component.} 

Once all the ESVMs in the ensemble have been trained, they all evaluate the candidate object proposals in new images: each ESVM produces a score, which can be interpreted as a similarity of the sample under evaluation to the positive proposal on which the ESVM has been trained on. Since it has been evaluated by every ESVM, the new sample receives as many scores as there are positive proposals in the training set. {We assign the proposal to the positive class ``animals'' if at least one ESVM has predicted this label (Fig.~\ref{fig:ESVM}b).}

In general, the scores produced by the different ESVMs cannot be compared directly, because the ESVMs have been trained independently to score the largest possible value on every single positive instance. As a consequence, each ESVM can score the same positive example with very different values, depending on the similarity of the candidate region to the example used for training. To ensure that the score provided by an ESVM on a test  {sample is} comparable to those of the other ESVMs, we normalize each score so that the distance between the margin (red line in Fig.~\ref{fig:ESVM}a) and the positive proposal is equal to one.

\subsection{Model improvement by active learning}\label{ssec:active_learning}

It is known that ground truths crowdsourced by querying non-experts are prone to inevitable errors introducing label noise, resulting in models that are harder to train and ultimately to lower accuracy~\citep{Hak10,Fon15}. To improve the quality of the ground truth and consequently the accuracy of the system specifically deal with two possible errors:

\begin{itemize}[noitemsep,nolistsep]
\item[-] \emph{False positives:} ground truth objects wrongly labeled as animals, while their correct class is ``background'';
\item[-] \emph{False negatives:} ground truth objects wrongly labeled as background, while their correct class is ``animal''.
\end{itemize}

On the one hand, false positives can be removed by a visual inspection of the proposals in the training set, since their number is limited (typically, we use 300 to 700 proposals in this study). On the other hand, false negatives cannot be found by systematic user inspection, since this set can easily contain tens of thousands of object proposals, and most of those will be of the actual background class. To tackle this task and lead the selection of a few negative examples to be screened by a user, we propose to use an iterative technique known as \emph{active learning}~\citep{tuia2011jstsp}. Active learning is based on a user-machine interaction: the machine asks the user to provide labels of the objects, for which the current prediction is not confident. Given the answer of the user, newly labeled examples are added to the training set {and the model becomes} more robust in areas of low confidence. 

Here, differently from standard active learning pipelines, we aim at finding wrongly labeled proposals in the training set, i.e. background proposals that contain actual animals, thus possibly wrongly annotated by annotators. By definition, these proposals have a visual appearance that is very similar to the ``animal'' class{, and consequently they lie close to the current EESVM decision boundary}.

\begin{figure}[!t]
\centering
\begin{tabular}{cc}
\includegraphics[width=.45\linewidth]{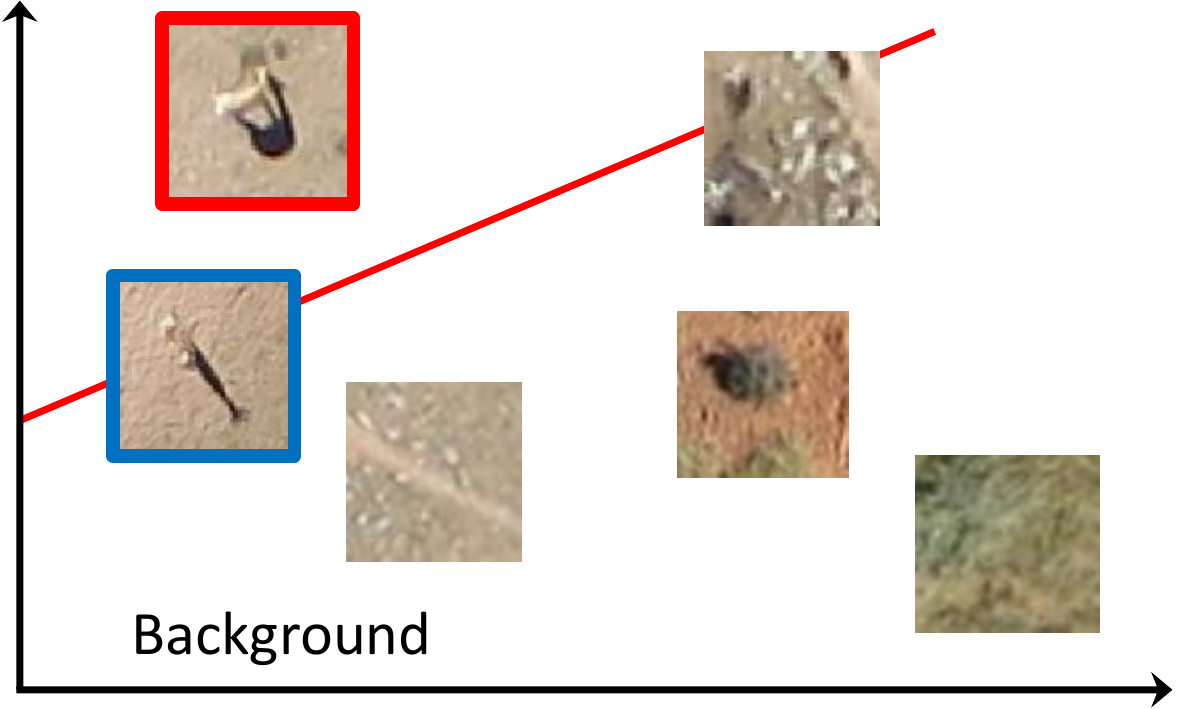}&
\includegraphics[width=.45\linewidth]{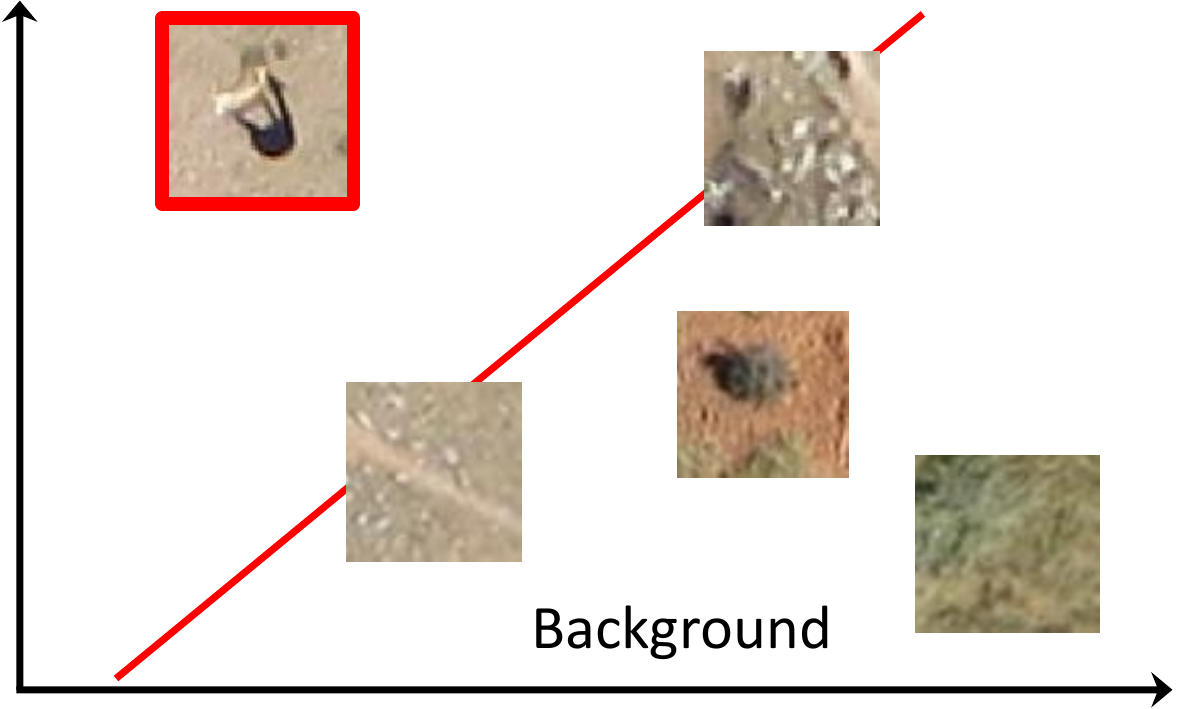}\\
(a) & (b)
\end{tabular}
\caption{Efficient search of false negative proposals with active learning. (a) The negative proposals with the highest score, i.e. the most similar to the positive proposal (framed in red in this example), are shown to the user, who recognizes that one of them (framed in blue) is a false negative. (b) The false negative is removed from the negative training set, thus modifying the decision function (red line) of the positive ESVM being trained. In addition, the false negative can be used as an additional ESVM in the model.}
\label{fig:ALflow}
\end{figure}

{We formulate our active learning routine as follows:} during training, an ESVM is trained on a single positive proposal and all the negative proposals. Then, this model assigns a detection score to all the negative objects in the training set. If a false negative similar to the exemplar is present, it will receive a high detection score. The top scoring objects are then shown to the user (Fig.~\ref{fig:ALflow}a), who is invited to inspect them via a graphic user interface. Following the user's response, three actions can be undertaken:
\begin{itemize}
\item[-] The proposal correctly belongs to the class ``background'': in this case, nothing happens and the proposal continues to be treated as a negative example in the training set. The next ESVM is trained normally.
\item[-] The proposal is a false negative: in this case the proposal is removed from the negative training set and the ESVM retrained without the confusing example (Fig.~\ref{fig:ALflow}b). The user can also choose to add the newly found animal to the positive training set, thus increasing the number of ESVMs by one;
\item[-] The user cannot decide: in case of extremely ambiguous proposals, we simply remove the proposal from the training set to ensure that no conflicting information is harming the learning process.
\end{itemize}

\section{Experiments and results} \label{sec:results}

In this section we present results on the Kuzikus dataset. First, we perform an evaluation on the impact of our hyperparametrization (Section~\ref{sec:par}). Then, we report the results obtained by our system after model selection, including the contribution of the active learning step (Section~\ref{sec:realres}).

\subsection{Ablation study: features used, their parameters and image resolution}\label{sec:par}

The ablation study evaluates the contribution of the different components of the proposed pipeline to the full model. To this end, we convert the animal detection problem into a balanced two-class classification problem and employ a linear SVM as base classifier. The reason behind this choice is that the analysis of the global factors of variation of the problem are much more robust when training models considering the whole class-conditional distribution, rather than depending from single positive examples in extremely unbalanced settings (EESVM). Furthermore, note that we employ linear SVM classifiers as main building blocks of our proposed EESVM.

The training and test sets comprise 1'324 and 568 proposals containing animals, respectively. {To compare to~\citet{ofli_combining_2016}, we use the same number of positive and negative examples}. To assess the random variability of our results, each experiment is repeated five times. Each run uses the same set of positive proposals, and {a randomly drawn negative set sampled over non-animal ground truth regions}. The hyperparameter trading off training errors and margin width of the linear SVM $C$ is selected via a 5-fold cross-validation. ROC curves averaged over the 5 runs report accuracy. Table~\ref{tab:exp_setup} summarizes the parameters considered in each experiment.

\begin{table}[!h]
\caption{Summary of the parameters considered in the experiments in Section~ \ref{sec:par}.}
\label{tab:exp_setup}
\centering
\begin{tabular}{rp{3cm}|c|p{3cm}|c}
\hline
Section & Experiment & GSD (cm) & Feature types & \# words \\\hline
\hline
\ref{ssec:ft}  & Feature types & 8 & HOC, BOVW, HOC+BOVW & 100\\\hline
\ref{ssec:nvw} & Number of visual words ($k$) & 8 & HOC+BOVW & 100, 300 \\\hline
\ref{ssec:ires}& Image resolution & 8, 12, 16 & HOC+BOVW & 100 \\
\hline
\end{tabular}
\end{table}

\subsubsection{Feature types}\label{ssec:ft}

\begin{figure}[!t]
\centering
\includegraphics[width=10cm]{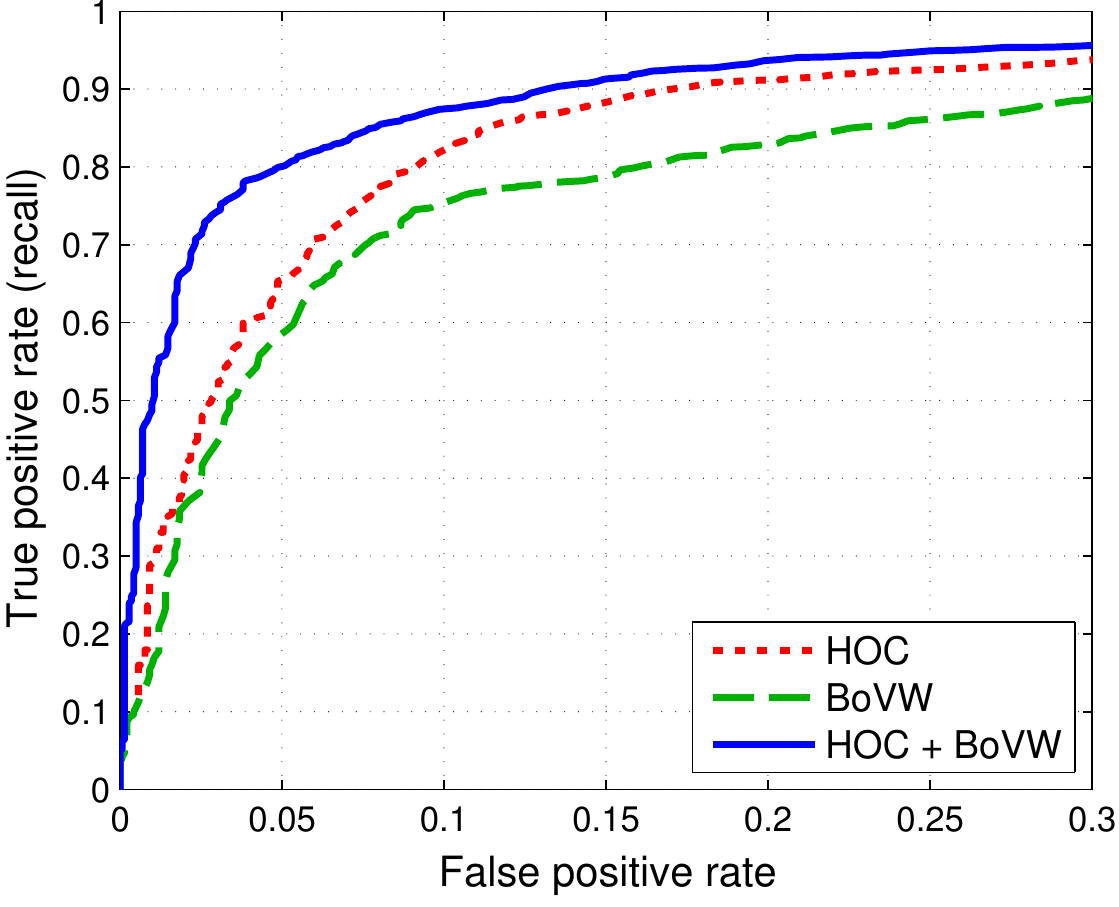}
\caption{Balanced classification scores per feature: HOC, BoVW and their concatenation.}
\label{fig:ROC_features_types}
\end{figure}

Figure~\ref{fig:ROC_features_types} presents the detection scores obtained with the HOC and BoVW features independently and with {their concatenation}. {To balance the relative importance of each feature type, the features are first normalized to $z$-scores. Then, they are further normalized to a unit $\ell_2$-norm, as suggested by \citep{kobayashi2015cvpr}.}

When used alone, the HOC features perform very well in comparison to the more elaborated and complex BoVW (with 100 words). This suggests that colors hold a large part of the relevant information, while the shapes and structures represented by the visual words seem to be less important. Both features perform similarly when requiring a small false positives rate. However, the combination of both features clearly improves over the single sets along the whole ROC curve. For instance, if a false positive rate of $0.03$ is retained, the recall for the combined features is $0.75$,  while being only $0.50$ for HOC and $0.45$ for BOVW. In the next experiments, we will use the concatenation of the feature types {to build the base appearance models}.

\subsubsection{Number of visual words -- $k$}\label{ssec:nvw}

\begin{figure}[!t]
\centering
\includegraphics[width=10cm]{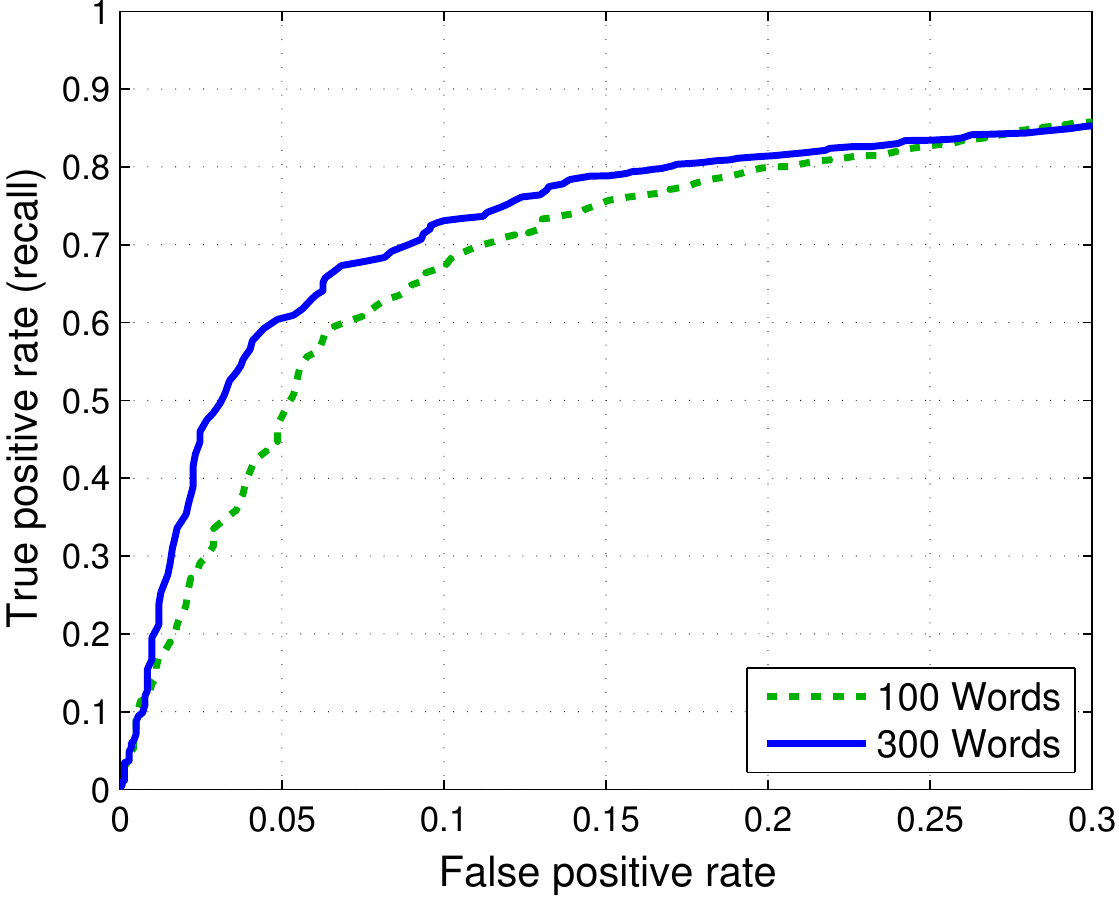}
\caption{Detection scores with 100 and 300 words.}\label{fig:ROC_number_words}
\end{figure}

Curves in Fig.~\ref{fig:ROC_number_words} illustrate how the number of visual words influences the performance of the models when trained on differently clustered BoVW features. Here, we show the effects for $k = 100$ and $k = 300$ words, when . Smaller {$k$ values} produced significantly worse results, while larger did not produce better accuracies. 

% Discussion number of visual words
As one could expect for problems involving complex appearance variations of positive and background classes, using more words improves the classification. The benefit is maximal for recall rates between $0.35$ and $0.60$. In this range, using $300$ words improves the recall rate up to 15\% (Fig.~\ref{fig:ROC_number_words}). The number of words required to properly describe the images content reveals the diversity and complexity of the patterns found in the dataset. While several thousands of visual words are often used with natural images, here a few hundred words can already retain most of the information.

\subsubsection{Image resolution}\label{ssec:ires}

\begin{figure}[!t]
\centering
\includegraphics[width=10cm]{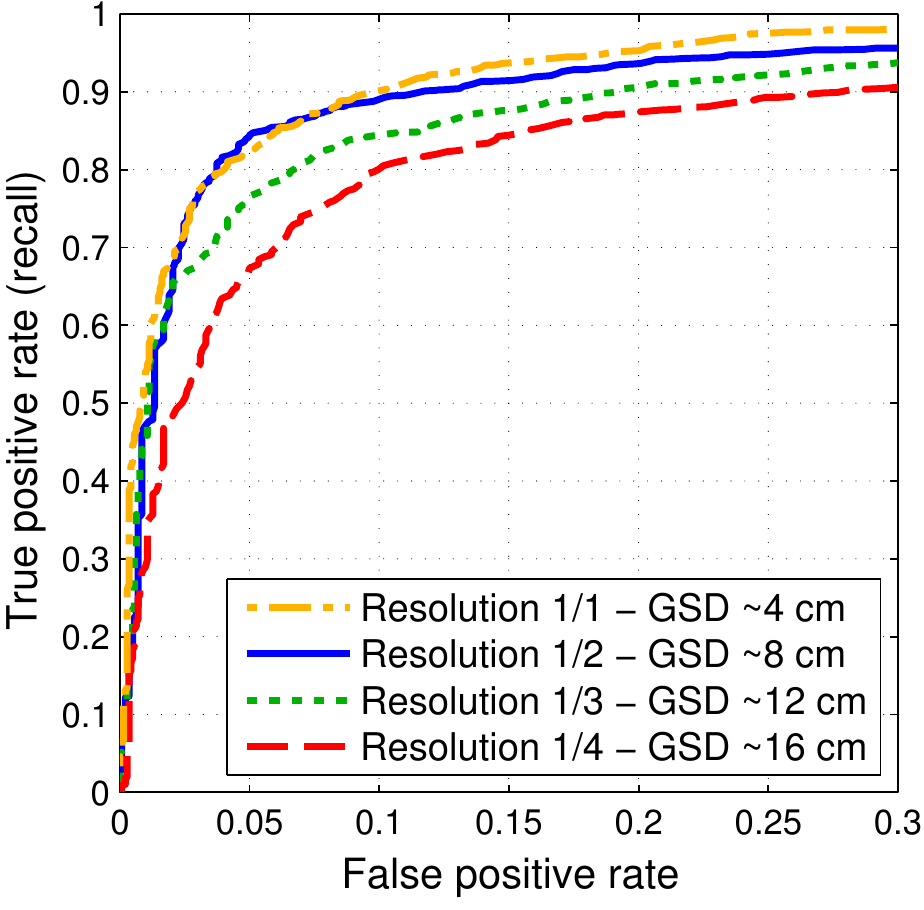}
\caption{Detection ROC curves with four different spatial resolutions.}\label{fig:ROC_resolution}
\end{figure}

The last ablation experiment concerns the spatial resolution of the images. Results in Fig.~\ref{fig:ROC_resolution} show the effect of reducing the original ground sampling distance (GSD), which is of approximatively 4 cm per pixel, to 8, 12 and 16 cm per pixel. Using the original image GSD did not improve the results significantly if comparing to a half resolution degradation, while it increased the computational time needed to extract features significantly. {Remember that the relation between the resolution and the computation for the BoVW descriptor is linear with a factor proportional to the increase in number of pixels per spatial unit.}

%Discussion image resolution
The curves suggest that a GSD of 16 cm is too coarse to detect animals. Interestingly, the benefits of using a GSD of 8 cm over a GSD of 12 cm only appear for recall rates of $0.65$. This indicates that two thirds of the animals do not require a GSD higher than 12cm, but the last third of the animals becomes more distinguishable when the resolution is {at least} of 8 cm. 

\subsection{Animals detection with ESVMs and unbalanced class-ratio}\label{sec:realres}

This section deals with the original task of detecting animals in the full dataset, characterized by a strongly unbalanced class-ratio. We employ models optimized thanks to the ablation study. From now on, all the \textit{background} objects are included in the training and test sets. Each of the original 654 annotated images (see Section~\ref{sec:data sets}) was assigned entirely to either the training or the test set, meaning that all the animals annotated on one image are included in the same set. This way, we avoid any spatial correlation between training and test sets. Both sets are supposed to contain a similar number of large, medium and isolated animals and to show similar probability distributions. The training set comprises 574 positive objects (animals) and 403'859 negatives, giving a ratio of 1:703. The test set has 284 positives and 160'384 negatives, yielding a ratio of 1:564.
 
\subsubsection{Results of the proposed system}

Figure~\ref{fig:PrecRec_active_learning} shows the precision-recall curves obtained with the proposed system (blue curve). Because animals are very rare in the dataset, achieving a high precision is difficult, but in our context it is more important to ensure a high recall. The false detections can be manually eliminated by the user in a further step, while it is much more difficult to recover animals missed by the detector. Our results show that indeed we can obtain high recall: for example we can achieve 75\% of correct detections for a precision of 10\%.
 
\begin{figure}[!t]
\begin{tabular}{cc|cc}
Detection & Most similar  &  Detection & Most similar \\
& proposal & & proposal\\
\includegraphics[width= .2\linewidth]{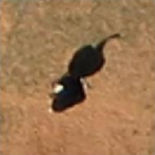} &
\includegraphics[width= .2\linewidth]{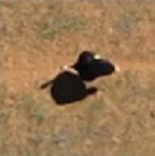} & 
\includegraphics[width= .2\linewidth]{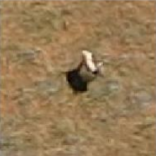} &
\includegraphics[width= .2\linewidth]{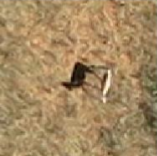}
\\
\multicolumn{2}{c|}{(a)} & \multicolumn{2}{c}{(b)} \\
\includegraphics[width= .2\linewidth]{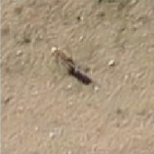} &
\includegraphics[width= .2\linewidth]{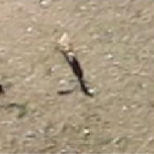} & 
\includegraphics[width= .2\linewidth]{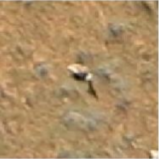} &
\includegraphics[width= .2\linewidth]{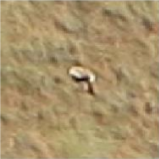}
\\
\multicolumn{2}{c|}{(c)} & \multicolumn{2}{c}{(d)} \\
\includegraphics[width= .2\linewidth]{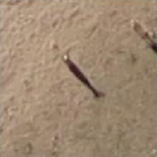} &
\includegraphics[width= .2\linewidth]{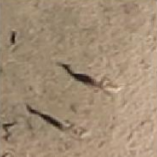} & 
\includegraphics[width= .2\linewidth]{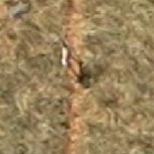} &
\includegraphics[width= .2\linewidth]{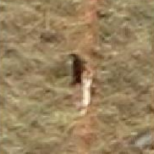}\\
\multicolumn{2}{c|}{(e)} & \multicolumn{2}{c}{(f)} \\
\end{tabular}
\caption{Detected animals (columns 1 and 3) and associated proposals (columns 2 and 4). The images are centred on the detections / proposals. Note that they are presented in full resolution and offer a larger view than the $25 \times 25$ pixels regions used for feature extraction.}\label{fig:captions}
\end{figure}

%Similarity between detections and exemplars, and metadata transfer
An interesting property of the EESVM is that a detection is always associated with the positive training example most similar to it. If additional information about the training examples is available (e.g. the species), it can be easily transferred to the detection directly at test time. Unfortunately, we could not quantitatively test such an idea for our dataset, because the ground truth did not include species information (in the crowdsourcing campaign only the presence/absence of animals was recorded). Nevertheless, we observe that many detections are visually very similar to their closest proposal, as illustrated in Fig.~\ref{fig:captions}. The appearance of the detection is not simply a direct matching of the proposal itself, but each ESVM learns the color statistics independently from the spatial orientation of the tiles: for example the detection depicted in Fig.~\ref{fig:captions}e shows animals in very different positions, while in Fig.~\ref{fig:captions}f the shadowing is reversed.

\subsubsection{Effect of the time of the day}

Here, we study the robustness of the detection pipeline with respect to the acquisition time of the day. The time of the day strongly influences the presence of shadows and the spectral response of the camera: models trained on morning images could be suboptimal for the detection of animals on images acquired in afternoon, since image statistics are different. This problem is known as domain adaptation~\citep{Tui15b}.

\begin{figure}[!t]
\begin{tabular}{cc}
\includegraphics[width= .48\linewidth]{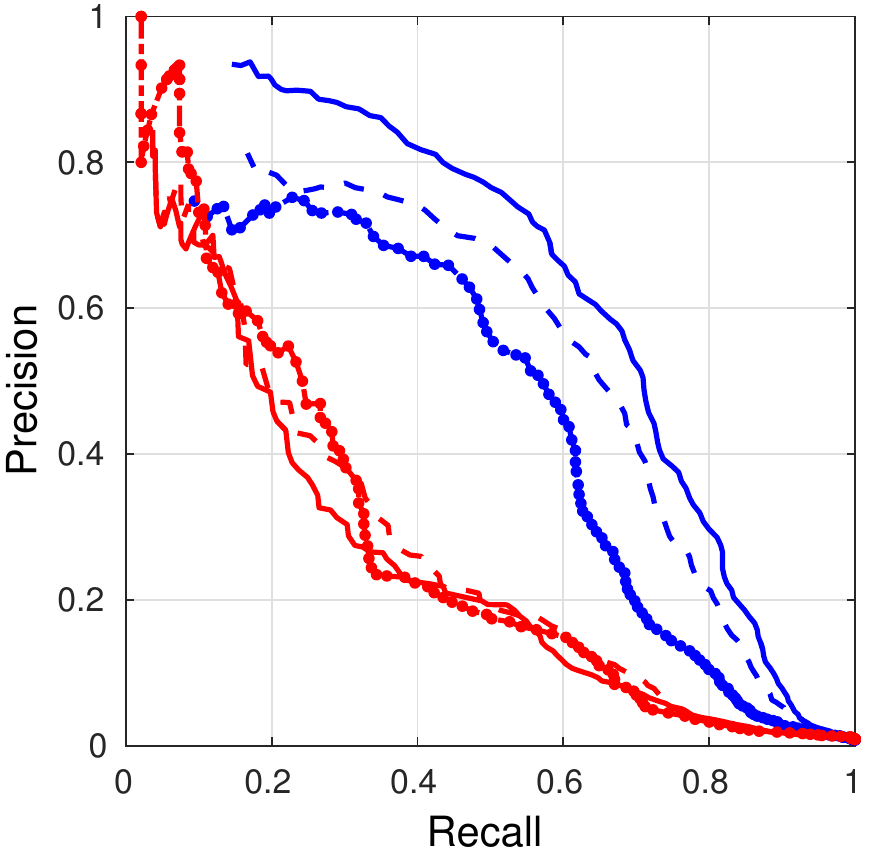} &
\raisebox{3cm}{\includegraphics[width= .48\linewidth]{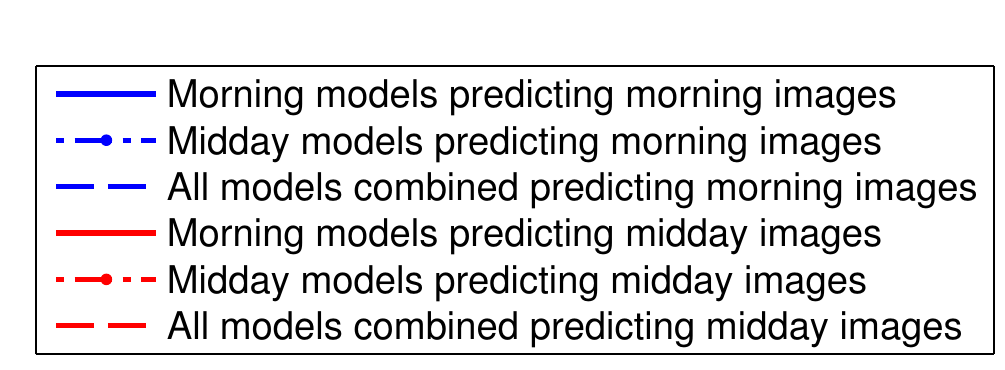}} \\
(a) & \\
\includegraphics[width= .48\linewidth]{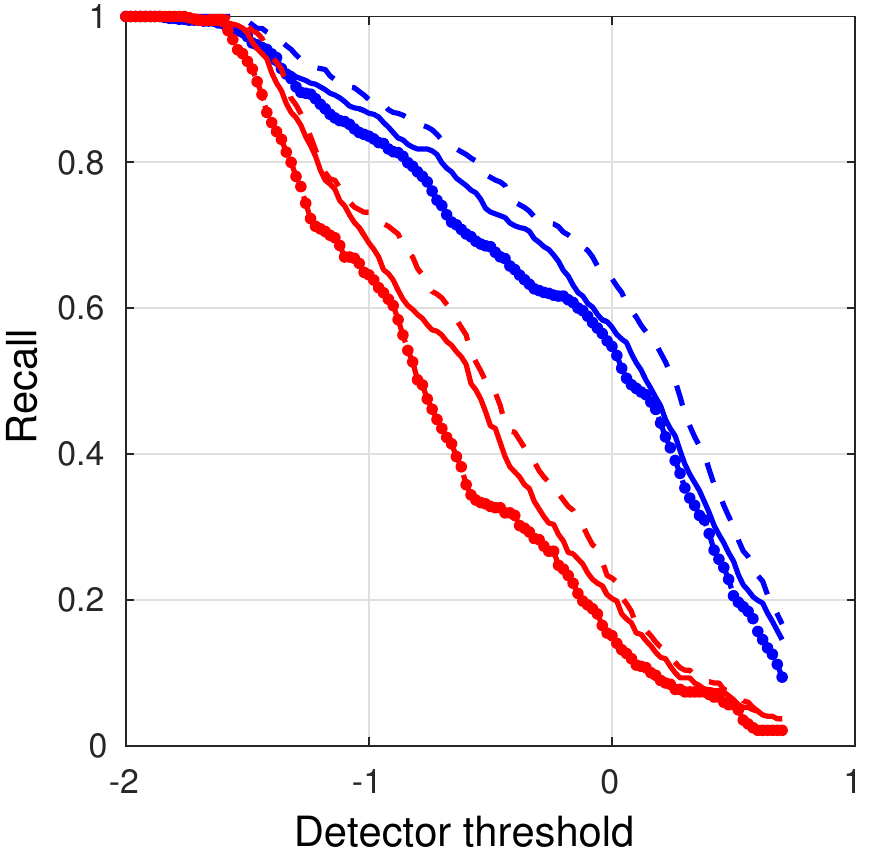} &
\includegraphics[width= .48\linewidth]{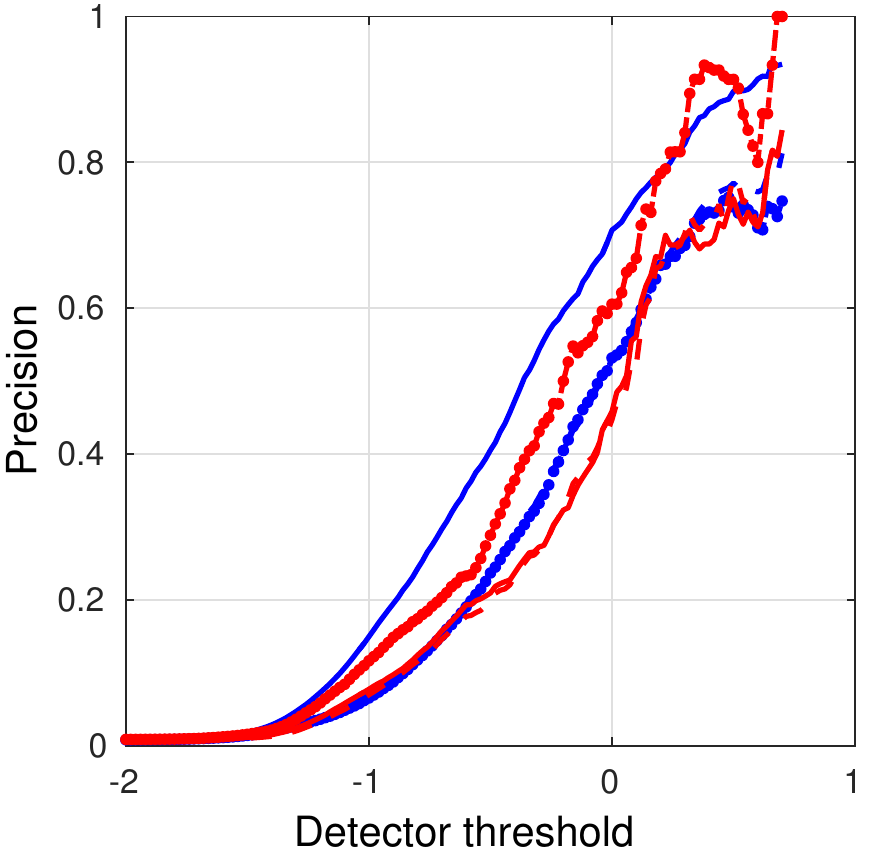} \\
(b) & (c)\\
\end{tabular}
\caption{Detection results {obtained by models applied on morning acquisition (red lines) and at noon (blue lines), respectively}. Dashed lines refer to {models trained} with proposals from the morning image set, dotted lines of the midday image set, and solid lines of both sets combined. (a) precision-recall curve; (b) recall and (c) precision curves as function of the detector threshold.} \label{fig:hour_full}
\end{figure}

To determine whether the time of the day influences the detection rate, we define two sets: one made of images taken in the morning (from 09h13 to 09h28) and another of images taken around midday (from 13h08 to 13h30), respectively. Each set is then subdivided in a training and a test subset, like in the experiments above. Table~\ref{tab:hour_subsets} indicates the number of images and animals in each subset. To ensure a fair comparison, we used $176$ positive examples for both time steps, corresponding to the total number of animals in the midday acquisitions. The detection results are reported in Fig.~\ref{fig:hour_full}:

\begin{table}[!t]
\centering
\caption{Composition of the subsets ``Morning'' and ``Midday''.}
\label{tab:hour_subsets}
\begin{tabular}{c|cc|cc}
& \multicolumn{2}{c|}{Training} & \multicolumn{2}{c}{Test}\\
\hline
& \# images & \# animals & \# images & \# animals\\
\hline
Morning & 89 & 176 & 36 & 119\\
\hline
Midday & 120 & 176 & 48 & 82\\
\hline
\end{tabular}
\end{table}

%Discussion time of the day
\begin{itemize}
\item[-] From the precision-recall curves (Fig.~\ref{fig:hour_full}a), we observe that the morning test set is easier to classify than the midday test set, regardless of the proposals involved in training. {We hypothesize that the more discriminative shadows play a role in this difference.}
\item[-] The recall curves (Fig.~\ref{fig:hour_full}b) consider the question `\emph{Given a detector threshold and all the animals in the test set, how many were correctly identified as positive detections?}'.
These curves indicate that models trained on morning data allow training more accurate models, even when such models are used to classify the midday test set. However, this result is always outperformed by the situation where training sets from both times have been jointly used, since the training set is larger and covering more variations in appearance.
\item[-] The precision rates (Fig.~\ref{fig:hour_full}c) consider the question `\emph{Given a detector threshold and all the positive detections, how many were real animals?}'. In this case, we observe another behaviour: the best results are obtained when the training and the test subsets are from the same time period. Similar results are observed for both the morning and midday datasets. On the contrary, mixing the acquisition times in training leads to a degradation of the results, mostly due to the many false positives. {These curves underline that in order to minimize the number of false positives, the images used to train the models must be acquired at a time as close as possible as the one when one wants to detect animals.}
\end{itemize}

We conclude that flying in the morning and always at the same hour of the day can lead to better results. However, this analysis may be biased by the fact that the morning and midday image sets were not taken over the exact same locations. Even thought the land cover is very similar, it is possible that one of the image sets contains more confusing objects or hiding places for the animals.

\subsubsection{Active learning}

\begin{figure}[!t]
\centering
\includegraphics[width=10cm]{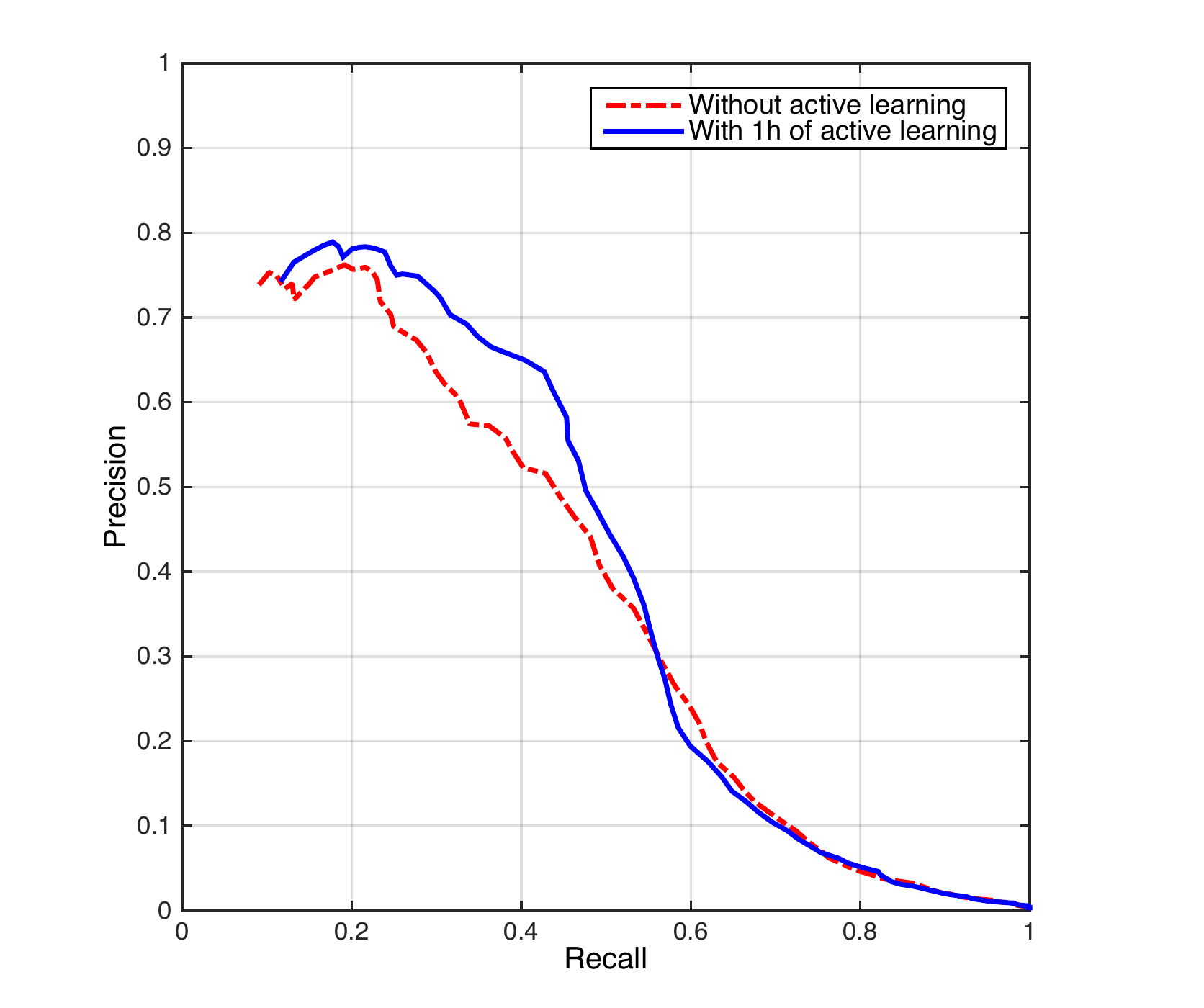}
\caption{Precision-recall curves of the classification without active learning ({red dashed line}) and with one hour of active learning ({blue solid line}).}\label{fig:PrecRec_active_learning}
\end{figure}

\begin{figure}[!t]
\includegraphics[width= 1\linewidth]{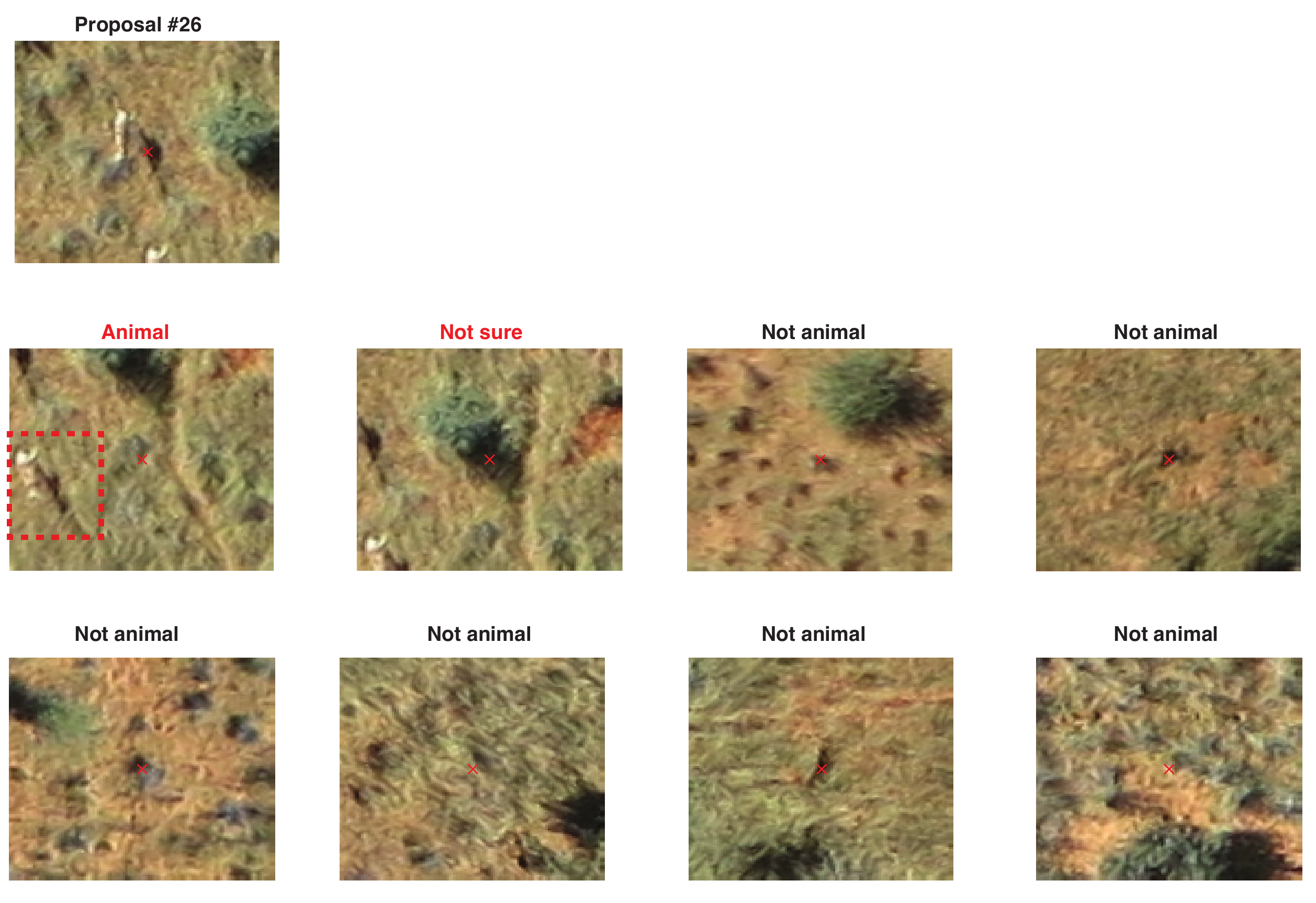}
\caption{Human computer interaction results for positive proposal \#$26$ (shown in the top-left panel). The eight top-scoring negative proposals are presented to the user (midlle and bottom row). The user could highlight one proposal including an animal (red-dashed box, highlighted for clarity), even if not centered. The user also decided to remove the second proposal, most probably because of the shadowing and a white shape in the bottom left corner. The other proposals are true negatives and can be left in the training set.}
\label{fig:AL}
\end{figure}

In this section, we study the possibility to refine the training data using active learning. In the previous section, we considered an ensemble of 574 ESVMs, {each one corresponding to a positive proposal, trained against} 403'859 negative {proposals}. We now aim at highlighting proposals of the negative set that can potentially contain an animal (\emph{false negatives}) and have them screened by a human user. We proceed sequentially {one ESVM at a time: a real user is asked to provide feedback on the eight most uncertain negative examples provided by the given model (Fig.~\ref{fig:AL}).}

In one hour, the user screened $120$ models. Among all negative proposals, $55$ were marked as animals and added to the positive set, thus raising the number of ESVM to $629$. In parallel, the user also marked $52$ originally negative proposals as unclear: these were simply removed from the set of negative proposals. After one hour, hardly any false negative {could be} found by the user. The detection results obtained with the $629$ ESVM is compared to the original one (obtained with $574$) in Fig.~\ref{fig:PrecRec_active_learning}. {Note that these results cannot be compared directly with those of Fig.~\ref{fig:hour_full}, as in this case we use a complete test set including images acquired both in the morning and at noon.}

%Discussion active learning
The precision-recall curves reveal that for this dataset, active learning  enhances the predictive ability of the EESVM for recall rates below 55\%. It does not help to find animals that are either difficult to detect (e.g. species that have never been seen by the system at training time), or observed in drastically different conditions. This form of sampling is an effective way to improve an existing ground truth: $55$ additional animals were found in the training images and we were able to remove proposals, for which a human expert was unable to decide whether an animal was present or not. This way, {a single user was} able to screen the entire negative set (with more than 400'000 proposals) within an hour. The user prioritized low-confidence one, thus showing the interest of using active learning instead of a random sampling strategy.

\section{Conclusions} \label{sec:conclusions}

In this paper, we proposed a {semi-}automated data-driven system to detect large mammals in the Semi-arid African Savanna. Such approaches are crucial to make the difference in near real-time conservation and war against wildlife poaching. Our system first processes many sub-decimeter UAV images to highlight possible candidate regions likely to contain animals (or \emph{proposals}), and then infer the presence of animals {among them by means of an object recognition model, the \emph{ensemble of exemplar SVM} (EESVM)}.

We study and discuss the impact of every system components by performing a complete ablation study, and highlight differences in the data representation (i.e. features) and other crucial aspects such as image resolution and acquisition time. For the purpose of detecting mammals, a resolution of $8$~cm proved to be sufficient when combined with the powerful histogram of colors and bag-of-visual-words descriptors.

When applied to the full problem, the proposed system achieves promising results and demonstrates that the detection of animals in aerial images in the semi-arid Savanna is feasible when employing simple RGB camera mounted on a UAV. Even if a high recall rate can be obtained, a human operator is required to verify the false negatives and to improve the available ground truth, a step that relies on active learning. Since our model is based on object proposals, it is also computationally advantageous over na\"ive techniques, as we only probe windows candidates likely to contain an animal. Furthermore, using the object proposal strategy jointly with the EESVM model opens for fine grained classification applications, such as the identification of animal species.

{Since it relies on static images acquired over a pre-defined flight-path, the current system is not able to provide exact counts for two main reasons. First, the same animal can be observed in more than one image with no way of disambiguating the detections. One option could be to plan image acquisition when animals are less active, but then also less visible. Instead, a promising idea would be to use UAV videos for making the detector aware of the temporal dimension: considering temporal trajectories makes it possible to re-identify animals based on their paths and more realistic counts can be provided.}

{The second reason is that the current system might detect two individuals that are very close as a single instance: in this case, one could use prior knowledge about the animal size to post-process the detections and possibly disambiguate animal clusters, for example by estimating size-constrained bounding boxes, of a size equal to the one of the animal being detected. However, although the approach presented here is able to only approximate the actual number of animals, it provides much more realistic numbers compared to traditional techniques. Finally, UAV data allow to process large geographical areas and the system presented in this paper is likely to represent a significant saving in money and time for wildlife land managers. It also represents a safe way to carry out animal surveys.}

\section*{Acknowledgements}

This work has been supported by the Swiss National Science Foundation (grant PZ00P2-136827 (DT, http://p3.snf.ch/project-136827). The authors would like to acknowledge the SAVMAP consortium (in particular Dr. Friedrich Reinhard of Kuzikus Wildlife Reserve, Namibia) and the QCRI and Micromappers (in particular Dr. Ferda Ofli and Ji Kim Lucas) for the support in the collection of ground truth data.

\section{References}

\bibliographystyle{elsarticle-harv}
\biboptions{authoryear}

%\bibliography{animalBib}
%\bibliography{PDMzotero}

\end{document}